\documentclass{article}

\usepackage[preprint, nonatbib]{neurips_2024}

\usepackage[utf8]{inputenc} 
\usepackage[T1]{fontenc}    
\usepackage{url}            
\usepackage{booktabs}       
\usepackage{amsfonts}       
\usepackage{nicefrac}       
\usepackage{microtype}      
\usepackage{xcolor}         
\usepackage{makecell}
\usepackage{adjustbox}
\usepackage{pifont}
\usepackage{bbding}
\usepackage{tabulary,multirow,xspace}
\usepackage{fixmath,mathtools,nicefrac,mmstyle}
\usepackage{subcaption}
\usepackage{caption}
\usepackage{float}
\usepackage{wrapfig} 
\usepackage[misc]{ifsym} 
\usepackage{colortbl}
\definecolor{mygray1}{gray}{.95}
\definecolor{mygray2}{gray}{.9}
\definecolor{mygray3}{gray}{.95}
\usepackage{pifont}
\newlength\savewidth
\newcolumntype{x}[1]{>{\centering\arraybackslash}p{#1pt}}

\newcommand{\app}{\raise.17ex\hbox{$\scriptstyle\sim$}}

\usepackage[pagebackref=true,breaklinks,colorlinks,citecolor=blue,linkcolor=blue,urlcolor=blue,hypertexnames=false]{hyperref}

\newcommand \footnoteONLYtext[1]
{
	\let \mybackup \thefootnote
	\let \thefootnote \relax
	\footnotetext{#1}
	\let \thefootnote \mybackup
	\let \mybackup \imareallyundefinedcommand
}

\title{Mamba or RWKV: Exploring High-Quality and High-Efficiency Segment Anything Model}

\author{ Haobo Yuan$^{1}$, Xiangtai Li$^{1, 2}$ \textsuperscript{$\dagger$}, Lu Qi$^{3}$, Tao Zhang$^{2}$, \\
\textbf{Ming-Hsuan Yang$^{3}$ Shuicheng Yan$^{2}$, Chen Change Loy$^{1}$} \\
  {$^{1}$S-Lab, NTU} 
  {$^{2}$Skywork AI} 
  {$^{3}$UC Merced}
  \\
  \textbf{Code: \url{https://github.com/HarborYuan/ovsam}}
}

\begin{document}

\maketitle

\begin{abstract}
\footnoteONLYtext{\textsuperscript{$\dagger$}: Project Lead. E-mail: xiangtai94@gmai.com and whuyuanhaobo@gmail.com.}
Transformer-based segmentation methods face the challenge of efficient inference when dealing with high-resolution images. 
Recently, several linear attention architectures, such as Mamba and RWKV, have attracted much attention as they can process long sequences efficiently.
In this work, we focus on designing an efficient segment-anything model by exploring these different architectures. 
Specifically, we design a mixed backbone that contains convolution and RWKV operation, which achieves the best for both accuracy and efficiency. 
In addition, we design an efficient decoder to utilize the multiscale tokens to obtain high-quality masks. 
We denote our method as RWKV-SAM, a simple, effective, fast baseline for SAM-like models.
%
Moreover, we build a benchmark containing various high-quality segmentation datasets and jointly train one efficient yet high-quality segmentation model using this benchmark.
%
%
Based on the benchmark results, our RWKV-SAM achieves outstanding performance in efficiency and segmentation quality compared to transformers and other linear attention models.
For example, compared with the same-scale transformer model, RWKV-SAM achieves more than 2$\times$ speedup and can achieve better segmentation performance on various datasets. In addition, RWKV-SAM outperforms recent vision Mamba models with better classification and semantic segmentation results.
Code and models will be publicly available. 
\end{abstract}

\section{Introduction}
\label{sec:intro}

Trained on large-scale segmentation datasets, Segment Anything Model (SAM)~\cite{kirillov2023segment} has recently garnered significant attention due to its remarkable versatility and effectiveness across numerous segmentation tasks. 
By taking visual prompts such as points and boxes provided by humans or other models as inputs, SAM can generate masks in various scenes, enabling various downstream applications such as image editing~\cite{Gao2023EditAnythingEU}, remote sensing~\cite{Hetang2024SegmentAM}, medical image segmentation~\cite{Ma2024SegmentAI}, etc. Despite its robust generalization capabilities, SAM exhibits several drawbacks that may hinder its practical applications in some scenarios. First, the computational cost of SAM is exceptionally high. Second, the segmentation quality of SAM still falls short in some cases; for example, SAM always generates overly smooth edges, which do not fit many cases. The above two drawbacks limit the application of SAM in real-time scenarios and fields requiring high-quality segmentation results.

Existing works usually only focus on solving either the first problem or the second problem. For example, several works~\cite{zhou2023edgesam,mobile_sam,xiong2023efficientsam}, such as EdgeSAM~\cite{zhou2023edgesam} and Efficient SAM~\cite{xiong2023efficientsam}, aim to explore efficient architecture for SAM. However, the segmentation quality is still limited. On the other hand, there are several works~\cite{sam_hq,song2024ba} explore high-resolution and high-quality SAM. They bring extra computational costs to SAM, which slows down the inference. Thus, a balance between high quality and high efficiency should be explored to better deploy SAM in real-world applications.

Recently, a series of works starting from the natural language processing community (e.g., RWKV~\cite{peng2023rwkv}, Mamba~\cite{gu2023mamba}) and following in the computer vision community (e.g., VMamba~\cite{liu2024vmamba}, Vision-RWKV~\cite{duan2024vrwkv}) have begun to focus on designing methods capable of handling long-range dependencies in linear time (linear attention models). 
Compared with transformers, where the complexity of their computation increases quadratically with the sequence length, the linear attention models reformulate the attention mechanism so that it scales linearly with the sequence length, thus significantly reducing computational costs when the sequence is very long. 
As a result, linear attention models can handle very long sequences while maintaining their global perception capability.
However, there are no previous works exploring these architectures on SAM-like promptable segmentation tasks.

\begin{figure}[t]
\centering
\includegraphics[width=1.\textwidth]{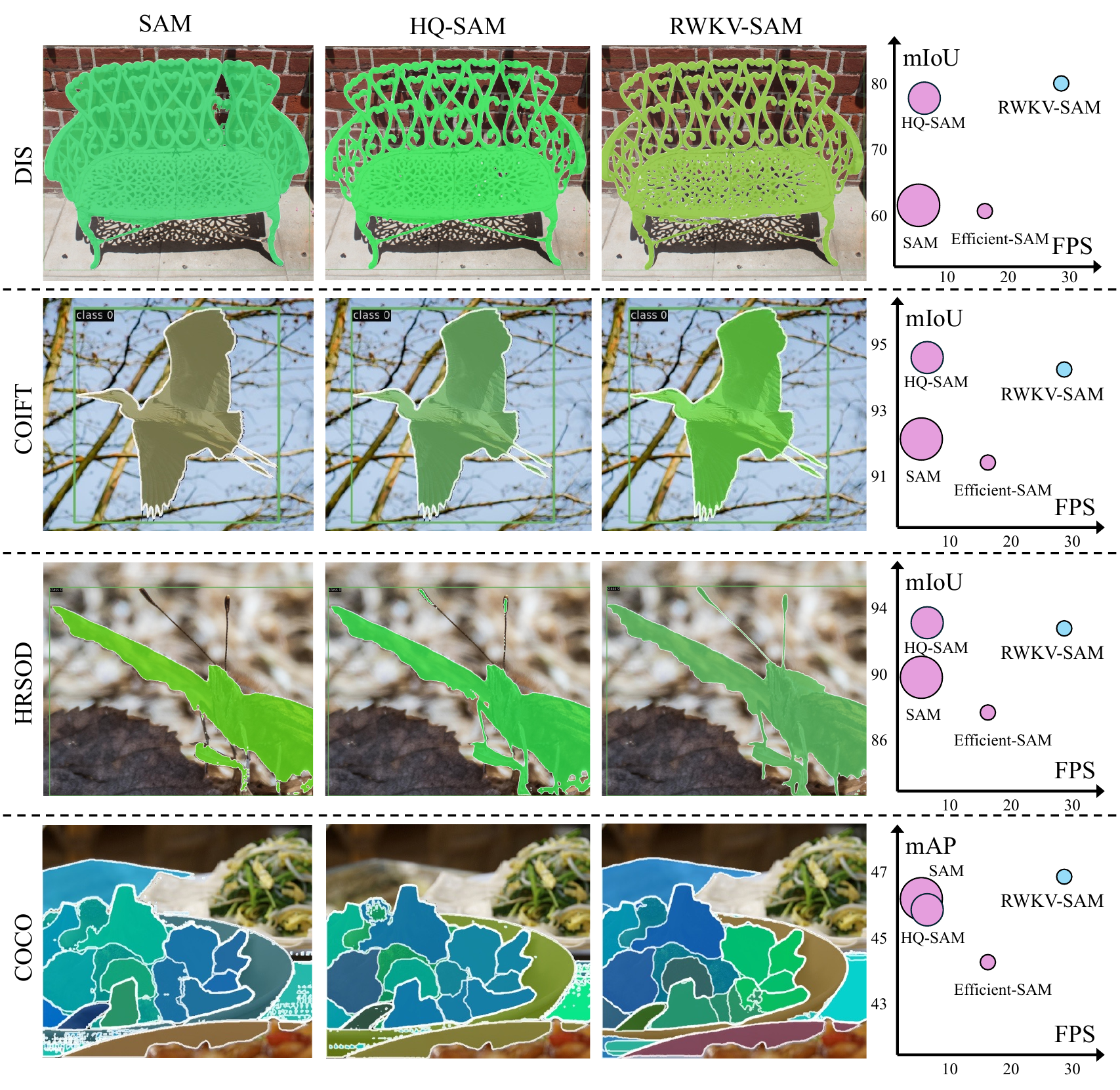}
\vspace{-5mm}
\caption{(Left) Comparisons between SAM~\cite{kirillov2023segment}, EfficientSAM~\cite{xiong2023efficientsam}, and our RWKV-SAM. (Right) FPS, parameters, and high-quality segmentation quality comparison of SAM~\cite{kirillov2023segment}, EfficientSAM~\cite{xiong2023efficientsam}, HQ-SAM~\cite{sam_hq}, and RWKV-SAM. The input image resolution is $1024 \times 1024$. We report the FPS and number of parameters of the backbone on one NVIDIA A100 GPU.}
\vspace{-5mm}
\label{fig:teaser}
\end{figure}

In this work, we try to solve these problems together to build an efficient and high-quality SAM using recent linear attention models.
In particular, we propose RWKV-SAM to handle SAM's computational cost and segmentation quality problems. 
The high computational cost of SAM can be attributed to two reasons: 1). extensive parameter count, and 2). quadratic time complexity caused by 
attention design in transformer layers as the input feature size grows.  
While prior efforts tackle the efficiency issue of SAM by reducing the model size (e.g., EfficientSAM~\cite{xiong2023efficientsam}), these solutions still face quadratic time complexity, which means they cannot achieve good efficiency in high-resolution inputs, for example, with $1024 \times 1024$ high-resolution inputs.
We propose an efficient segmentation backbone leveraging the RWKV~\cite{peng2023rwkv} to improve the efficiency in the high-resolution while maintaining the global perception. 
Our efficient segmentation backbone contains three stages, which the decoder can use to refine the generated masks. 
In addition, we explore different decoder designs to fuse the different scales of the features and train the model on a combined high-quality dataset to enable our RWKV-SAM as a high-quality and high-efficiency segment-anything model.

We evaluate our method on various datasets and benchmarks. As depicted in Figure~\ref{fig:teaser}, our RWKV-SAM outperforms previous methods in efficiency and quality.
Although it only requires about 1/16 of the inference time of SAM, our method achieves more accurate, high-quality segmentation results. Even though the model size is comparable to EfficientSAM, our RWKV-SAM runs more than 2x faster. Our method performs better than HQ-SAM with greater detail due to the information from low-level local features from the backbone.
Compared with previous linear models, such as Mamba, our RWKV-SAM runs faster and performs better on various benchmarks. 
Our model runs even faster when using extremely high-resolution inputs. 
We have the following contributions to this work: (1) We propose RWKV-SAM, which contains an efficient segmentation backbone that yields different resolutions of feature maps and leverages the RWKV operation to reduce time complexity. 
(2) We explore different designs to leverage the multiscale feature maps in the decoder and train the RWKV-SAM on the high-quality segmentation datasets to enable the high-quality segmentation capability. 
(3) We demonstrate the effectiveness of RWKV-SAM on several benchmarks, surpassing previous methods while maintaining efficiency. 
(4) We conduct detailed comparison studies on various linear attention models, including Vision Mamba and VRWKV. To our knowledge, this is the first work to explore these models in a fair comparison manner.
\section{Related Work}
\label{sec:related_work}

\noindent
\textbf{Efficient Segmentation.} Existing methods~\cite{ICnet,SFnet,bisenet,hu2023you,hong2021lpsnet,wan2023seaformer,yu2021bisenetv2,mehta2018espnet} on efficient segmentation have mainly concentrated on closed-set and specific domains~\cite{li2023transformer}. 
Much of the efficient segmentation research~\cite{SFnet,hu2023you,espnetv2} is dedicated to driving scenarios.
Also, multiple studies have been conducted on efficient panoptic segmentation~\cite{sun2023remax,hu2023you,mohan2020efficientps} and fast video instance segmentation~\cite{CrossVIS,zhang2023mobileinst}. 
Recently, various studies~\cite{zhang2022topformer, wan2023seaformer,zhou2023edgesam,zhou2023context,xu2024rapsam} have developed efficient segmentation techniques that facilitate model execution on mobile devices for the segment anything model. 
Mobile SAM~\cite{mobile_sam} introduces a streamlined encoder distillation method. Fast SAM~\cite{zhao2023fastsam} employs a single-stage instance segmentation framework that directly decodes class-agnostic masks. Edge SAM~\cite{zhou2023edgesam} deploys the SAM model on a real-world mobile device with a new prompt-guided distillation.
Efficient SAM~\cite{xiong2023efficientsam} 
In this work, in addition to the real-time constraint, we also aim for high-quality segmentation.

\noindent
\textbf{Efficient Backbone.} This direction primarily concentrates on developing efficient CNNs~\cite{mnetv1,mnetv2,mnetv3,squeezenet,shufflenetv1,shufflenetv2,ghostnet}, transformers~\cite{edgenext,edgevit,mvitv1}, and hybrid architectures~\cite{EMOiccv23,mocovit,mobileformer,nextvit,zhou2022transvod}, to learn visual representations. 
Recently, several works~\cite{duan2024vrwkv,liu2024vmamba,li2024videomamba,zhu2024vision,zhang2024point} have explored linear attention models, including RWKV~\cite{peng2023rwkv} and Mamba~\cite{gu2023mamba} in vision~\cite{zhu2024vision, duan2024vrwkv}. However, all these works try to replace the transformer for representation learning, ignoring the generating features at different scales. We explore an efficient backbone for fast, high-resolution segmentation, where we adopt the CNN and RWKV mixed architecture. Based on the experiments, our proposed backbone achieves better representation in similar parameters and latency constraints.

\noindent
\textbf{High-Quality Segmentation.} Previous works for high-quality segmentation aim for specific tasks via designing specific modules~\cite{kirillov2020pointrend, transfiner,li2020improving,shen2022high}, proposing fine-grained datasets~\cite{qi2022fine,qi2022open}, focusing on object-centric settings~\cite{qin2022,liew2021deep,han2022slim}, and adding refiner~\cite{cheng2020cascadepsp,SegRefiner}. To allow more open settings, several works~\cite{sam_hq,liu2024rethinking,song2024ba} have explored SAM as a base model to improve the segmentation quality. However, they cannot run in real time. In particular, HQ-SAM~\cite{sam_hq} brings extra costs compared to the original SAM. 
We have two goals compared with these works. One is to design a new model to segment high-quality object masks in real time. The other is to build an entire training pipeline, including datasets, to enable an efficient model for high-quality segmentation.

\noindent
\textbf{Linear Attention Models.} The transformer has computation cost issues when the token numbers become larger, which is exactly the challenge faced by high-quality segmentation. Recently, several works~\cite{shen2021efficient,qin2024hgrn2,yang2023gated,qin2024hierarchically} have shown great potential to replace transformer architecture. In particular, state space models~\cite{gu2023mamba,gu2022efficiently} have been proven to model long-range dependency. Moreover, RWKV~\cite{peng2023rwkv, peng2024eagle} is another method with faster inference speed. We aim to explore these architectures for efficient, high-quality segmentation under the segment anything model meta-architecture. In particular, we find that under the efficient segmentation setting of high-resolution image inputs, RWKV runs faster than Mamba. Thus, we aim to explore RWKV architecture as our backbone.

\section{Method}

\label{sec:method}
\textbf{Overview.} We aim to build an efficient, high-quality segment-anything model. That requires the model to have the following properties. \textbf{First}, the model should have a backbone that is efficient even in the high-resolution. \textbf{Second}, the model should be able to utilize existing SAM knowledge to avoid training on the whole SA-1B~\cite{kirillov2023segment}. \textbf{Third}, the model should be able to utilize the feature pyramid from the backbone and be trained using high-quality data to generate high-quality masks. To build a model fulfilling the three properties, we design an RWKV-based backbone (Section~\ref{sec:method_1}), which has a feature pyramid and is efficient in the high-resolution while having good performance compared to other transformer or linear attention models (please refer to Table~\ref{tab:main_results} and Figure~\ref{fig:Efficiency_analysis}). In Section~\ref{sec:method_2}, we introduce our training pipeline to use knowledge from the SAM model and high-quality datasets. We also present the decoder to fuse the features from different resolutions.

\begin{figure}[t]
\centering
\includegraphics[width=1.\textwidth]{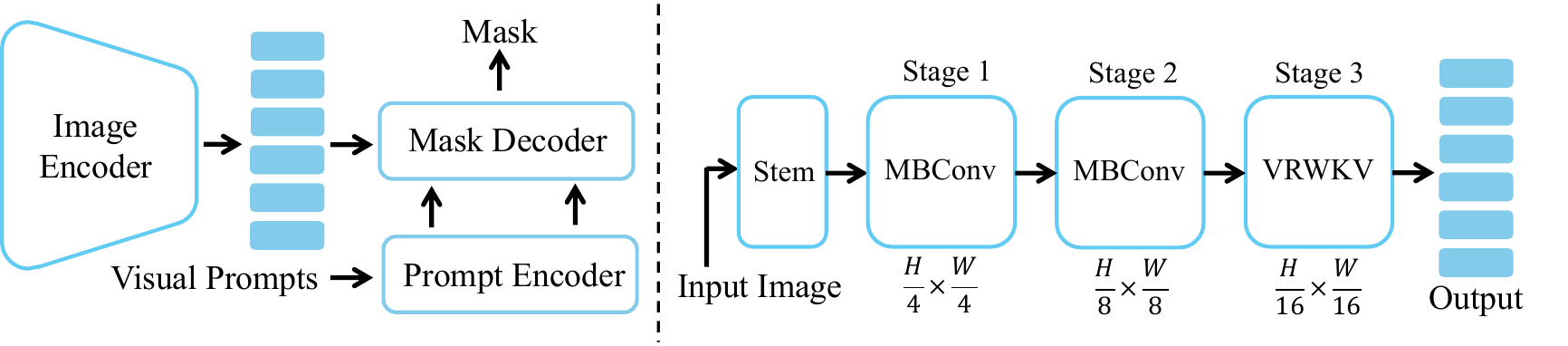}
\vspace{-5mm}
\caption{(Left) Overview of our RWKV-SAM. RWKV-SAM contains an image encoder, a prompt encoder, and a mask decoder. (Right) The efficient segmentation backbone architecture. The first two stages use the MBConv blocks, and the third uses the VRWKV blocks.}
\vspace{-5mm}
\label{fig:backbone}
\end{figure}

\subsection{Efficient Segmentation Backbone}\label{sec:method_1}
The original Segment Anything Model~\cite{kirillov2023segment} adopts a transformer-based backbone. Although it achieves powerful performance, it has a huge computational overhead. Although EfficientSAM~\cite{xiong2023efficientsam} reduces the number of model parameters drastically, it does not change the vision transformer architecture and still requires a long inference time at high resolution. The main reason for that lies in the intrinsic property of the vision transformer architecture. As the resolution increases, the number of patches grows quadratic, leading to increased computational demands.

To build an efficient segmentation backbone at high resolution, we follow the spirit of linear-time sequence modeling in the NLP community~\cite{gu2023mamba,peng2023rwkv} and propose an RWKV-based efficient vision backbone. In general, our backbone has a 3-stage design and contains two types of blocks: Mobile Convolution Blocks (MBConv)~\cite{mnetv2} and Vision-RWKV Blocks (VRWKV)~\cite{duan2024vrwkv}.

\begin{wraptable}{r}{0.52\textwidth}
 \vspace{-4mm}
 \renewcommand\tabcolsep{2pt}
 \renewcommand\arraystretch{1.1}
 \resizebox{\linewidth}{!}{
     \begin{tabular}{cc|cc|cc|cc} 
        \toprule
          &&\multicolumn{2}{c|}{\small{R-SAM-T}} & \multicolumn{2}{c|}{\small{R-SAM-S}} & \multicolumn{2}{c}{\small{R-SAM-B}}\\
          \small{Stage} & \small{Stride} & \small{\#block} & \small{\#chn} & \small{\#block} & \small{\#chn} & \small{\#block} & \small{\#chn}\\
         \midrule
         1 & 4 & 2 & 32 & 2 & 64 & 2 & 128 \\
         2 & 8 & 4 & 64 & 4 & 128 & 4 & 256\\
         3 & 16 & 14 & 192 & 14 & 384 & 14 & 768\\
         \midrule
         \multicolumn{2}{c|}{\#Param} & \multicolumn{2}{c|}{5.0M} & \multicolumn{2}{c|}{19.7M} & \multicolumn{2}{c}{78.7M} \\
        \bottomrule
     \end{tabular}
 }
 \vspace{-1mm}
 \caption{\small{Settings of different variants of backbone.}}
 \label{tab:backbone}
 \vspace{-3mm}
\end{wraptable}
\textbf{Macro-Level Design.} Figure~\ref{fig:backbone} shows the overview of our efficient segmentation backbone. 
The macro-level design of our backbone is motivated by ViTamin~\cite{chen2024vitamin}. In the first two stages, we employ the conv-based blocks, i.e., MBConv, to generate high-resolution feature maps. We downsample 2$\times$ the feature maps before each stage. The high-resolution feature maps can be used for mask refinement in the decoder. Before the third stage, the feature maps have a downsampling factor of 16, which means each pixel in the feature maps can be seen as a ``token''. In the third stage, we stack a series of VRWKV blocks, taking the tokens as input. Compared to plain vision transformer or Vision-RWKV~\cite{duan2024vrwkv}, our backbone has different scales of feature maps rather than a single fixed resolution. We present the settings of different variants of our backbone in Table~\ref{tab:backbone}. These multi-scale feature maps allow our model to adaptively focus on various spatial details, enhancing its ability to handle complex scenes with different object sizes.

\textbf{Micro-Level Design.} The MBConv block uses the ``inverted bottleneck'' design~\cite{mnetv2}. It contains a 1 × 1 convolution to expand the channel size, a 3 × 3 depthwise convolution for spatial mixing, and another 1 x 1 convolution to project the channel back to the original channel size. Following ViTamin~\cite{chen2024vitamin}, we use LayerNorm rather than BatchNorm for simplicity. We set the expand ratio to 4 in the MBConv block.

For the VRWKV block, the tokens are first processed by the spatial-mix module and then fed into the channel-mix module. The spatial-mix module serves as the role of global perception. Supposing the input tokens can be represented as $X\in\mathbb{R}^{L\times C}$, where $L$ indicates the token length and $C$ indicates the channel size, the spatial-mix module starts with the ${\rm Q\mbox{-}Shift}$ modules:
\begin{equation}
        R_\text{s} = {\rm Q\mbox{-}Shift}_R(X) W_R,~~
        K_\text{s} = {\rm Q\mbox{-}Shift}_K(X) W_K ,~~
        V_\text{s} = {\rm Q\mbox{-}Shift}_V(X) W_V . \\
\end{equation}
The ${\rm Q\mbox{-}Shift}$ is an important module since it allows each token to interpolate with 4-direction pixel neighborhoods, maintaining the locality of image features, and 
has been shown to be effective~\cite{duan2024vrwkv}:
\begin{equation}
    {\rm Q\mbox{-}Shift}(X) = X + (1 - \mu)X',
\end{equation}
where $X'$ is the token obtained by combining the four pixels around each token along the channel dimension. $\mu$ is a learnable scalar that is different for each representation.

After mixing the pixel neighborhoods, the spatial-mix module fuses tokens globally and bidirectionally:
\begin{equation}
    O_\text{s} = (\sigma (R_\text{s}) \odot \mathrm{Bi\mbox{-}WKV}(K_\text{s}, V_\text{s}))W_O,
\end{equation}
where $\sigma$ indicates the sigmoid function, $\odot$ is the element-wise multiplication. The $\mathrm{Bi\mbox{-}WKV}$ is the key component of the ``attention'' mechanism that allows each token to interact globally with all other tokens in the sequence. For each token at index $t$ in the sequence, with the $K_\text{s}\in\mathbb{R}^{L\times C}$ and $V_\text{s}\in\mathbb{R}^{L\times C}$ as input, it can be calculated as follows:
\begin{equation}
    \begin{aligned}
        \mathrm{Bi\mbox{-}WKV}(K,V)_t=\frac{\sum^{L-1}_{i=0,i\neq t}e^{-(|t-i|-1)/L \cdot w + k_i }v_i + e^{u + k_t}v_t}{\sum^{L-1}_{i=0,i\neq t}e^{-(|t-i|-1)/L \cdot w + k_i} + e^{u + k_t}},
    \end{aligned}
\end{equation}
where $w$ and $u$ are parameters shared globally in the sequence, and $k_i$ and $v_i$ corresponds to the feature $K_\text{s}$ and $V_\text{s}$ at index $i$.
The $\mathrm{Bi\mbox{-}WKV}(K,V)$ can be converted to RNN-Form to be executed within linear computational complexity and in parallel. Please refer to Vision-RWKV~\cite{duan2024vrwkv} for details. After the spatial-mix module, the tokens are fed into the channel-mix module:
\begin{equation}
\begin{aligned}
    R_\text{c} = {\rm Q\mbox{-}Shift}_R(X) W_R,~~
    K_\text{c} = {\rm Q\mbox{-}Shift}_K(X) W_K ,\\
    O_\text{c} = (\sigma (R_\text{c}) \odot \mathrm{SquaredReLU}(K_\text{c})W_V)W_O.
\end{aligned}
\end{equation}
The channel-mix module is calculated independently for each token, similar to MLP, but it adds ${\rm Q\mbox{-}Shift}$ to maintain the image feature locality further. In particular, $W_K$ projects the embedding to expand the embedding by two times, and $W_V$ projects the embedding to the original size.

\subsection{RWKV-SAM: Data, Model, and Training Pipeline}\label{sec:method_2}
\textbf{SAM Revisited.} The original Segment Anything Model (SAM) contains a heavy ViT-H~\cite{VIT} backbone, a prompt encoder that takes boxes or points as visual prompts, and a lightweight decoder that contains two transformer layers on the 16x downsampling stride. The lightweight decoder takes the backbone output and prompt encoder and generates the corresponding masks. SAM is trained on the large-scale auto-labeled SA-1B dataset containing 11M images, which requires 256 A100 GPUs for 68 hours. To build an efficient, high-quality segment anything model, our RWKV-SAM involves the design of training data, model structure, and training pipeline. 

\begin{wraptable}{r}{0.36\textwidth}
 \vspace{-4mm}
 \renewcommand\tabcolsep{2.0pt}
 \renewcommand\arraystretch{1.1}
 \resizebox{\linewidth}{!}{
     \begin{tabular}{lcc} 
        \toprule
         Datasets & \#Images & \#Masks \\
         \midrule
         COCONut-B~\cite{deng2024coconut} & 242K & 2.78M\\
         EntitySeg~\cite{qi2022fine} & 30k & 579k  \\
         DIS5K~\cite{qin2022} & 3k & 3k\\
         \bottomrule
     \end{tabular}
 }
 \vspace{-1mm}
 \caption{\small{Datasets for training.}}
 \label{tab:datasets}
 \vspace{-3mm}
\end{wraptable}
\textbf{Training Data.}
The annotations of the SA-1B~\cite{kirillov2023segment} used by SAM are generated automatically. 
Although it helps scale the training data, the annotations do not contain details.  
To mitigate the gap, we introduce three heterogeneous datasets for joint training. 
The first dataset is the COCONut-B~\cite{deng2024coconut}, which has 242K images, including the COCO~\cite{coco_dataset} labeled and unlabelled images. 
COCONut-B's annotations are generated by an assisted manual annotation pipeline, which yields high-quality annotations beyond the original annotations. 
The second dataset is EntitySeg~\cite{qi2022fine}, which contains 30k high-resolution (2000px to 8000px) images annotated by human annotators. 
The third dataset is DIS5K~\cite{qin2022} dataset. DIS5K dataset provides remarkable single-object, highly accurate annotations. 

\textbf{Model.} To generate accurate masks, the rich semantic context and low-level boundary details are both important. 
The macro-level design of our efficient segmentation backbone contains three stages. 
The output of the first two stages can be used as the low-level local features (4x and 8x downsampling stride), and the output of the third stage (16x downsampling stride) can be used as the global feature. We denote the features from the first two stages as $X_\text{hr}$ and $X_\text{mr}$, and the output of the third stage as $X$. We keep the prompt encoder ($\Phi_\text{pe}$) and decoder ($\Phi_\text{dec}$) for saving the knowledge from the original SAM. The original SAM takes the output of $\Phi_\text{pe}$ and $X$ as inputs to generate mask features $F_\text{M}$:
\begin{equation}
    F_\text{M} = \Phi_\text{dec}(\Phi_\text{pe}(P), X),
\end{equation}
where $P$ is the visual prompts. To further refine the mask features with low-level local features, introduce additional refine module $\Phi'_\text{dec}$ to incorporate $X_\text{hr}$ and $X_\text{mr}$:
\begin{equation}
    F'_\text{M} = \Phi'_\text{dec}(F_\text{M}, X, X_\text{mr}, X_\text{hr}),
\end{equation}
where $F'_\text{M}$ is the refined mask features. We explore several designs of $\Phi'_\text{dec}$ and use two convolution layers to fuse features for simplicity and efficiency. The refined mask features can be used to generate the mask outputs $M = Q {\otimes} F'_\text{M}$, where $Q$ is the instance query generated by $\Phi_\text{dec}$ and $\otimes$ represents the dot product for each mask.

\textbf{Training Pipeline.} Our RWKV-SAM consists of a two-step training process. In the first step, we employ the original SAM (VIT-H) to distill our efficient segmentation backbone. 
We follow Open-Vocabulary SAM~\cite{yuan2024ovsam} to use a per-pixel mean squared error (MSE) loss for aligning the efficient segmentation backbone to the VIT-H backbone:
\begin{equation}
\begin{aligned}
    L_{\text{S}1} = \mathrm{MSE}(X_{\text{SAM}}, X),
\end{aligned}
\end{equation}
where $X$ is the output of the RWKV-SAM backbone and the $X_{\text{SAM}}$ is the output of the VIT-H backbone in SAM. 
In the second step, we utilize the combined datasets to conduct joint training of the whole model. For each image, we first generate the bounding box of each instance based on the mask annotation and randomly select up to 20 instances for training. After generating masks by RWKV-SAM based on the visual prompts, we apply mask Cross Entropy (CE) loss and Dice loss~\cite{milletari2016v} between the ground truth masks and the generated masks. The loss of the second step can be formulated as:
\begin{equation}
    L_{\text{S}2} = \lambda_{\text{ce}}L_{\text{ce}} + \lambda_{\text{dice}}L_{\text{dice}}.
\end{equation}
We follow previous works~\cite{cheng2021mask2former, cheng2021maskformer} to set both $\lambda_{\text{ce}}$ and $\lambda_{\text{dice}}$ to 5.

\section{Experiments}
\label{sec:exp}

\begin{table}
 \renewcommand\tabcolsep{3.5pt}
 \renewcommand\arraystretch{1.2}
 \resizebox{\linewidth}{!}{
     \begin{tabular}{l|cc|cc|cc|cc|cc} 
        \toprule
         \multirow{2}{*}{Method} & \multicolumn{2}{c|}{Stat} & \multicolumn{2}{c|}{COCO~\cite{coco_dataset}} & \multicolumn{2}{c|}{DIS~\cite{qin2022}} & \multicolumn{2}{c|}{COIFT~\cite{liew2021deep}} & \multicolumn{2}{c}{HRSOD~\cite{Zeng2019TowardsHS}} \\
            & \#Param & FPS &  mAP & mBAP & mIoU & mBIoU & mIoU & mBIoU & mIoU & mBIoU \\
         \midrule
         SAM (ViT-H)~\cite{kirillov2023segment} & 600M & 2.5 & 46.1 & 31.3 & 62.0 & 52.8 & 92.1 & 86.5 & 90.2 & 83.1 \\
          EfficientSAM~\cite{xiong2023efficientsam} & 22.4M & 17.8 & 44.4 & 30.1 & 61.3 &	52.0&	91.4&	85.5&	88.2&	80.8 \\
         HQ-SAM~\cite{sam_hq} & 300M & 4.4 & 46.6 & 31.7 &  78.6&70.4&94.8	&90.1&	93.6&86.9 \\
        \midrule
        VMamba-S~\cite{liu2024vmamba} + SAM & 49.4M & 34.7 & 45.7 & 31.2 & 81.2 &	76.0 &	93.2 &87.4 &	92.7&	86.8 \\
        VRWKV-S~\cite{duan2024vrwkv} + SAM & 23.4M & 46.7& 45.2 & 30.2 & 75.9&69.2&92.8&86.4&90.6&	83.5  \\
        Vim-S~\cite{zhu2024vision} + SAM & 27.3M & 32.2 & 45.1 & 29.8 & 78.1 & 71.9 & 92.7 & 86.3 & 91.5 & 84.7 \\
        \hline
        RWKV-SAM (Ours) & 19.7M & 40.3 & 46.9 & 32.2 & 80.5 & 75.2 & 94.1& 88.7& 92.5& 86.5\\
         \bottomrule
     \end{tabular}
 }
 \vspace{1mm}
 \caption{SAM-like benchmarks. SAM~\cite{kirillov2023segment} and EfficientSAM~\cite{xiong2023efficientsam} are trained on the large-scale SA-1B datasets. HQ-SAM~\cite{sam_hq} is trained on their proposed HQ datasets. Other methods, including RWKV-SAM, are trained on the combination of COCONut-B~\cite{deng2024coconut}, EntitySeg~\cite{qi2022fine}, and DIS5K~\cite{qin2022}.}
 \label{tab:main_results}
\end{table}

\subsection{Experimental Setup}\label{sec:exp_1}
\textbf{Datasets.} Our training involves two sessions as mentioned Section~\ref{sec:method_2}. In the first session, we use 1\% of SA-1B data to distill the efficient segmentation backbone. We train the 1\% of SA-1B for 24 epochs (equivalent to 24 epochs on COCO). In the second session, we use the combination of COCONut-B~\cite{deng2024coconut}, EntitySeg~\cite{qi2022fine}, and DIS5K~\cite{qin2022} datasets for training. As the data samples differ among different datasets, we repeat the EntitySeg and DIS5K datasets to balance the three datasets to a 2:1:1 ratio, resulting in 482k samples for each epoch. We train the combined datasets for 6 epochs (equivalent to 24 epochs on COCO).

\textbf{Evaluation Protocol.} Since our method is trained on various high-quality datasets, including single-object and complex scenarios, we select three types of benchmarks for evaluation considering different scenarios. The first benchmark is on COCO datasets. We use a strong detector, ViTDet-H~\cite{Li2022ExploringPV}, to generate the bounding boxes as the visual prompts as the prompt inputs in the segment anything model. The second benchmark is on the DIS~\cite{qin2022} dataset (validation set). With the bounding box generated by the mask annotations as prompt inputs, we test the performance of generated masks on the single-object high-quality dataset. The third benchmark also has single-object datasets but includes COIFT~\cite{liew2021deep} and HR-SOD~\cite{Zeng2019TowardsHS} datasets to test the zero-shot performance on the single-object high-quality datasets. We report the mask mean AP (mAP) and mask boundary mean AP (mBAP) on COCO and report the mIoU and boundary mIoU (mBIoU) on single-object segmentation datasets.

\subsection{Main Results}

\begin{table}
 \vspace{-4mm}
 \centering
 \renewcommand\tabcolsep{3.5pt}
 \renewcommand\arraystretch{1.1}
 \resizebox{0.7\linewidth}{!}{
     \begin{tabular}{l|l|ccc} 
       \toprule
        Method & Block  & Emb Dim & \#Param &Accuracy\\
        \midrule
        VRWKV-T~\cite{duan2024vrwkv} & VRWKV & 192 & 6.0M & 75.1\\
        Vim-T~\cite{zhu2024vision} & Mamba & 192 & 7.9M & 76.1\\
        RWKV-SAM-T (Ours) & VRWKV + MBConv & 192  & 5.0M  & 75.6\\
        \midrule
        Vim-S~\cite{zhu2024vision} & Mamba & 384 & 27.3M & 80.5\\
        VRWKV-S~\cite{duan2024vrwkv} & VRWKV & 384 & 23.4M & 80.1\\
        RWKV-SAM-S (Ours) & VRWKV + MBConv & 384 & 19.7M & 80.9\\
        \midrule
        VRWKV-B~\cite{duan2024vrwkv} & VRWKV & 768 & 93.7M & 82.0\\
        RWKV-SAM-B (Ours) & RWKV + MBConv & 768 & 78.7M  & 82.5\\
        \bottomrule
     \end{tabular}
 }
 \vspace{1mm}
 \caption{Top-1 accuracy on ImageNet.}
 \vspace{-3mm}
 \label{tab:imagenet}
\end{table}
\textbf{ImageNet Pretraining.} Our proposed efficient segmentation backbone is first pertaining to the ImageNet-1K~\cite{russakovsky2015imagenet} dataset. We train the backbone 120 epochs with 224x224 resolution and using the receipt of Swin-Transformer~\cite{liu2021swin} following VRWKV~\cite{duan2024vrwkv}. To validate the performance of the backbone, we test the ImageNet classification performance on the validation set. As shown in Table~\ref{tab:imagenet}, with smaller parameter numbers, our method performs better or comparable compared to previous methods based on Mamba or RWKV.  For example, when comparing the small version of Vim~\cite{zhu2024vision}, VRWKV~\cite{duan2024vrwkv}, and RWKV-SAM, at the same embedding dimension 384, our method still outperforms previous methods despite having a smaller model size. 
We argue that the improvement may benefit from our macro-level design, which uses convolution layers in the first two blocks to obtain features at different scales instead of directly downsampling by transforming images to image patches.

\begin{wraptable}{r}{0.5\textwidth}
 \centering
 \renewcommand\tabcolsep{3.5pt}
 \renewcommand\arraystretch{1.1}
 \resizebox{\linewidth}{!}{
     \begin{tabular}{l|ccc} 
       \toprule
        Backbone & Decoder & \#Param &mIoU\\
        \midrule
        DeiT-T~\cite{VIT} & UperNet~\cite{xiao2018unified} & 5.7M & 39.2\\
        Vim-T~\cite{zhu2024vision} & UperNet~\cite{xiao2018unified} & 7.9M & 41.0\\
        RWKV-SAM-T (Ours) & UperNet~\cite{xiao2018unified} & 5.9M & 41.1\\
        \midrule
        DeiT-S~\cite{VIT} & UperNet~\cite{xiao2018unified} & 22.0M & 44.0\\
        Vim-S~\cite{zhu2024vision} & UperNet~\cite{xiao2018unified} & 27.3M & 44.9\\
        RWKV-SAM-S (Ours) & UperNet~\cite{xiao2018unified} & 23.6M & 45.3\\
        \bottomrule
     \end{tabular}
 }
 \caption{\small{Results of semantic segmentation on ADE20K. We report the number of parameters of the backbone.}}
 \label{tab:ade}
\end{wraptable}
\textbf{Semantic Segmentation.} To test the semantic segmentation performance of the backbone, we use the pre-trained backbone as the feature extractor and integrate a UperNet~\cite{xiao2018unified} as the decoder. 
We use the same setting to train the model for 160k iterations on the ADE20K dataset~\cite{ADE20K}. For the RWKV-SAM, We made minor modifications to the RWKV-SAM backbone, incorporating an MBConv block at the end of the third stage to generate features at the smallest scale (1/32). 
This feature, combined with the outputs from the first three blocks of the efficient segmentation backbone, forms a feature pyramid. 
As in Table~\ref{tab:ade}, we report the comparison results with Vim~\cite{zhu2024vision}. The results show that our method performs better even with fewer parameters.

\textbf{Segment Anything Model.} We compare our RWKV-SAM equipped with the efficient segmentation backbone with previous works on the benchmarks mentioned in Seciton~\ref{sec:exp_1}. As shown in Table~\ref{tab:main_results}, although SAM~\cite{kirillov2023segment} and EfficientSAM~\cite{xiong2023efficientsam} show good performance on the COCO dataset, it falls short compared to our method and HQ-SAM~\cite{sam_hq} in the high-quality datasets. 
On the COIFT~\cite{liew2021deep} and HR-SOD~\cite{Zeng2019TowardsHS} datasets, although our training dataset does not contain data samples in the same domain, our method still demonstrates strong zero-shot performance compared to the much larger SAM~\cite{kirillov2023segment}. 
We also train previous linear attention models with the segment anything decoder. VRWKV-S~\cite{duan2024vrwkv} also uses VRWKV blocks, but it only has one feature scale. 
Thus, it cannot use high-resolution features to optimize the segmentation results. 
Based on the results, it has a relatively worse performance on the high-quality datasets. Using the bi-directional mamba layer~\cite{gu2023mamba}, Vim~\cite{zhu2024vision} also has a relatively worse performance, especially on the boundary metric of the COCO dataset.

\subsection{Ablation Study and Analysis}
\begin{wrapfigure}{r}{0.44\textwidth}
    \vspace{-4mm}
    \centering
    \includegraphics[width=0.43\textwidth]{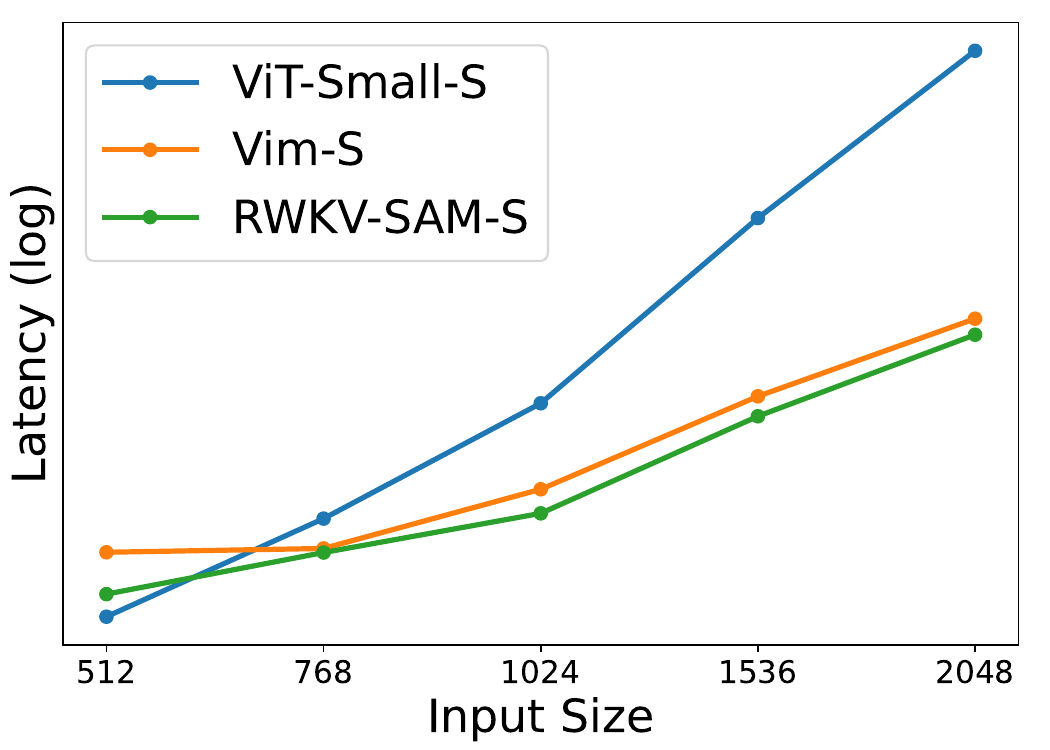}
    \caption{\small{Latency (log scaled) of the backbone with different input image resolutions.}}
    \label{fig:Efficiency_analysis}
\end{wrapfigure}
\textbf{Efficiency Analysis.} 
In Figure~\ref{fig:Efficiency_analysis}, we report the latency of the RWKV-SAM-Small and ViT-Small (used by EfficientSAM~\cite{xiong2023efficientsam}) with different input image sizes. The results are tested on a single NVIDIA A100 GPU. Both the two backbones adopt the 384 embedding dimension. With relatively low input resolutions, the FPS of the two models are similar (e.g., with 512x512 image as input, ViT-Small is 86.8 FPS and RWKV-SAM-Small is 73.4 FPS). 
However, when the resolution of the input image grows, the latency of the ViT model increases quadratically. In contrast, the latency of RWKV-SAM-Small grows linearly, which means it has advantages when the input image size is large. 
As shown in Figure~\ref{fig:Efficiency_analysis}, the latency of RWKV-SAM still maintains a significant advantage compared to ViT when using the high-resolution image as input. 
In the segment anything model, the typical input size is 1024x1024. 
Despite using the same input size, RWKV-SAM-Small achieves a significantly higher FPS (40.3) than ViT-Small (17.8) due to its more efficient computation mechanism. Consequently, using a similarly scaled model (ViT-Small: 22.4M, RWKV-SAM-S: 19.7M), RWKV-SAM has a better efficiency as the segment anything model backbone.

\begin{table}
 \renewcommand\tabcolsep{2pt}
 \renewcommand\arraystretch{1.2}
  \resizebox{0.58\linewidth}{!}{
     \begin{tabular}{l|c|ccc} 
       \toprule
        Method & FPS & DIS & COIFT & HRSOD\\
        \midrule
        RWKV-SAM + Conv Fusion & 32.5 & 80.5  & 94.1& 92.5\\
        RWKV-SAM + Mamba Fusion & 31.0 & 78.2 &	93.2 &	91.2\\
        RWKV-SAM + RWKV Fusion & 30.5 & 78.7 &	93.8 &	91.9 \\
        RWKV-SAM + DCN Fusion & 27.8 & 80.3&	93.9&	92.3\\
        \bottomrule
     \end{tabular}
 }
\hfill
 \resizebox{0.40\linewidth}{!}{
     \begin{tabular}{l|cccc} 
        \toprule
        Method & FPS & \#Param & Acc\\
        \midrule
        RWKVSAM-S & 40.3 & 19.7M  & 80.9\\
        RWKVSAM-S (1-1) & 35.9 & 17.2M &79.8\\
        RWKVSAM-S (2-2) & 32.6 & 14.8M & 78.6\\ 
        RWKVSAM-S (3-3) & 29.6 & 12.3M & 78.1\\
         \bottomrule
     \end{tabular}
 }\hfill
 \vspace{1mm}
 \caption{Alation studies. We test FPS results on NVIDIA A100 GPU. (Left) The ablation study of the decoder design. We report the FPS for the entire model, including the encoder and decoder, with one prompt as input. We report mIoU for comparison. (Right) The ablation study on the encoder design. (x-y) refers to the RWKV blocks used in the first two stages.}
 \label{tab:ablation}
\end{table}
\textbf{Effect of the Fusion Module in Decoder.} In the decoder, we explore the different designs to fuse the features from different scales. The first design uses two convolution layers for each scale to downsampling the low-level features and align the channels and the following two convolution layers after fusing the three features along the channel dimension. The second design replaces the convolution layers with RWKV block, which enables the global perception of the features. The third design is gradually fusing from low-resolution features to high-resolution and uses DCN for fusing following FaPN~\cite{huang2021fapn}. This design may give the model more opportunities to capture information from a relatively long range. 
As shown in Table~\ref{tab:ablation} (left), the first design has the best efficiency while achieving comparable performance compared to the others. The second design may hurt the performance, indicating that using RWKV block in the decoder is not as effective as using local operators such as convolution. 
We suspect this may be because RWKV blocks break the continuity of image features. 
Therefore, we use the first design mentioned in Section~\ref{sec:method_2}.

\textbf{Ablation of the Encoder Design.} As mentioned in Section~\ref{sec:method_2}, we use a 3-stage design. In the first two stages, we use the MBConv blocks to learn the low-level representations. We explore the effect of adding some RWKV blocks to the first stages to evaluate the macro-level design. In Table~\ref{tab:ablation} (right), we report the FPS (under 1024x1024 input size) and the accuracy of the alternative designs. 
There are a total of 14 RWKV blocks in the RWKV-SAM backbone, and they are in the third stage by default. In the table, (x-y) means that x and y blocks are put in the first two stages, and other blocks are in the third stage. Based on the results, with more blocks in the first two stages, although the model size is reduced (the hidden embedding size of the first two stages is smaller), the inference speed in the 1024$\times$1024 input size is slower. This design also leads to a decrease in model performance. Therefore, we put the RWKV blocks in the third stage for better inference speed.

\begin{figure}
    \centering
    \includegraphics[width=\linewidth]{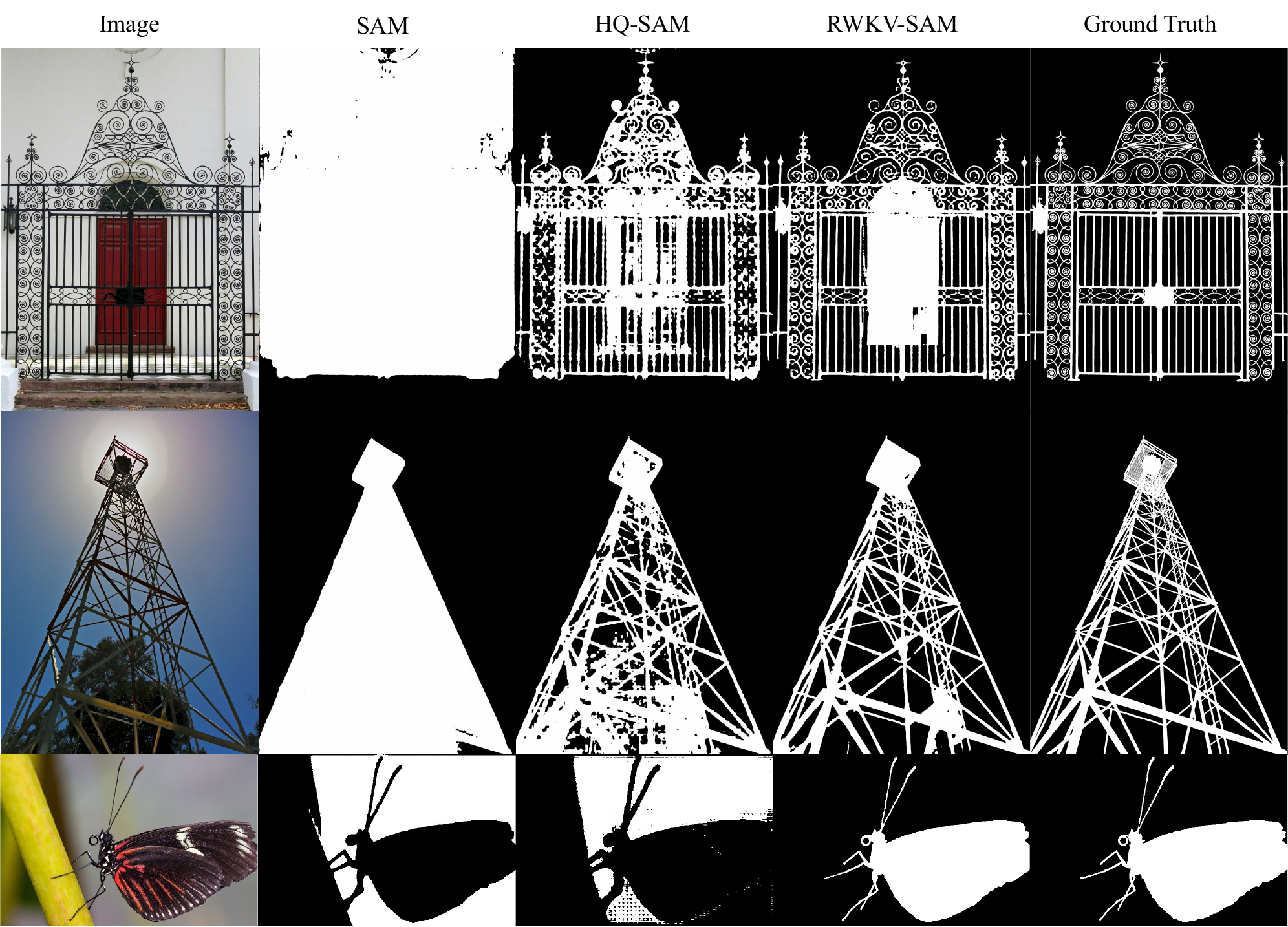}
    \caption{Visualization comparison. Given the box prompt as input, we show the predicted segmentation masks. Our RWKV-SAM shows better segmentation performance, especially in terms of detail.}
    \label{fig:vis}
    \vspace{-2mm}
\end{figure}
\textbf{Visualization.} 
In Figure~\ref{fig:vis}, we compare our method with SAM~\cite{kirillov2023segment} and HQ-SAM~\cite{sam_hq}. Based on the visualization results, we observe that our method achieves superior segmentation quality even in very complex scenes. For example, in the first example, SAM cannot distinguish the fence gate, and the HQ-SAM loses a lot of details. In contrast, our RWKV-SAM achieves the best results regarding the details. 
We show more visualization results in the appendix.

\section{Conclusion}
\label{sec:conclusion}
In this paper, we develop RWKV-SAM, which includes an efficient segmentation backbone and a complete training pipeline to enable the high-quality segmentation capability for segment anything model. 
Benefiting from the linear complexity, our method achieves excellent efficiency at high resolution while maintaining a strong performance. 
After training on the proposed benchmark, our RWKV-SAM demonstrates superior high-quality segmentation performance. 
We also benchmark recently proposed linear attention models, including Mamba and RWKV, showing that RWKV-SAM performs well among them. 
Our RWKV-SAM can segment any object with high quality and high efficiency, making it a generalized segmentation tool that can facilitate downstream applications.
We hope our research can inspire new architectural designs using linear attention models for dense prediction tasks.
{
  \bibliographystyle{plain}
  \bibliography{refbib}

\begin{thebibliography}{10}

\bibitem{chen2024vitamin}
Jieneng Chen, Qihang Yu, Xiaohui Shen, Alan Yuille, and Liang-Chieh Chen.
\newblock Vitamin: Designing {Scalable} {Vision} {Models} in the {Vision}-{Language} {Era}.
\newblock In {\em CVPR}, 2024.

\bibitem{mobileformer}
Yinpeng Chen, Xiyang Dai, Dongdong Chen, Mengchen Liu, Xiaoyi Dong, Lu~Yuan, and Zicheng Liu.
\newblock Mobile-former: Bridging mobilenet and transformer.
\newblock In {\em CVPR}, 2022.

\bibitem{cheng2021mask2former}
Bowen Cheng, Ishan Misra, Alexander~G. Schwing, Alexander Kirillov, and Rohit Girdhar.
\newblock Masked-attention mask transformer for universal image segmentation.
\newblock In {\em CVPR}, 2022.

\bibitem{cheng2021maskformer}
Bowen Cheng, Alexander~G. Schwing, and Alexander Kirillov.
\newblock Per-pixel classification is not all you need for semantic segmentation.
\newblock In {\em NeurIPS}, 2021.

\bibitem{cheng2020cascadepsp}
Ho~Kei Cheng, Jihoon Chung, Yu-Wing Tai, and Chi-Keung Tang.
\newblock {CascadePSP}: Toward class-agnostic and very high-resolution segmentation via global and local refinement.
\newblock In {\em CVPR}, 2020.

\bibitem{deng2024coconut}
Xueqing Deng, Qihang Yu, Peng Wang, Xiaohui Shen, and Liang-Chieh Chen.
\newblock Coconut: Modernizing {COCO} {Segmentation}.
\newblock In {\em CVPR}, 2024.

\bibitem{VIT}
Alexey Dosovitskiy, Lucas Beyer, Alexander Kolesnikov, Dirk Weissenborn, Xiaohua Zhai, Thomas Unterthiner, Mostafa Dehghani, Matthias Minderer, Georg Heigold, Sylvain Gelly, et~al.
\newblock An image is worth 16x16 words: Transformers for image recognition at scale.
\newblock {\em ICLR}, 2021.

\bibitem{duan2024vrwkv}
Yuchen Duan, Weiyun Wang, Zhe Chen, Xizhou Zhu, Lewei Lu, Tong Lu, Yu~Qiao, Hongsheng Li, Jifeng Dai, and Wenhai Wang.
\newblock Vision-rwkv: Efficient and scalable visual perception with rwkv-like architectures.
\newblock {\em arXiv preprint arXiv:2403.02308}, 2024.

\bibitem{Gao2023EditAnythingEU}
Shanghua Gao, Zhijie Lin, Xingyu Xie, Pan Zhou, Ming-Ming Cheng, and Shuicheng Yan.
\newblock Editanything: Empowering unparalleled flexibility in image editing and generation.
\newblock In {\em ACM MM}, 2023.

\bibitem{gu2023mamba}
Albert Gu and Tri Dao.
\newblock Mamba: Linear-time sequence modeling with selective state spaces.
\newblock {\em arXiv preprint arXiv:2312.00752}, 2023.

\bibitem{gu2022efficiently}
Albert Gu, Karan Goel, and Christopher R\'e.
\newblock Efficiently modeling long sequences with structured state spaces.
\newblock In {\em ICLR}, 2022.

\bibitem{ghostnet}
Kai Han, Yunhe Wang, Qi~Tian, Jianyuan Guo, Chunjing Xu, and Chang Xu.
\newblock Ghostnet: More features from cheap operations.
\newblock In {\em CVPR}, 2020.

\bibitem{han2022slim}
Kunyang Han, Jun~Hao Liew, Jiashi Feng, Huawei Tian, Yao Zhao, and Yunchao Wei.
\newblock Slim scissors: Segmenting thin object from synthetic background.
\newblock In {\em ECCV}, 2022.

\bibitem{Hetang2024SegmentAM}
Congrui Hetang, Haoru Xue, Cindy~X. Le, Tianwei Yue, Wenping Wang, and Yihui He.
\newblock Segment anything model for road network graph extraction.
\newblock In {\em CVPRW}, 2024.

\bibitem{hong2021lpsnet}
Weixiang Hong, Qingpei Guo, Wei Zhang, Jingdong Chen, and Wei Chu.
\newblock Lpsnet: A lightweight solution for fast panoptic segmentation.
\newblock In {\em CVPR}, 2021.

\bibitem{mnetv3}
Andrew Howard, Mark Sandler, Grace Chu, Liang-Chieh Chen, Bo~Chen, Mingxing Tan, Weijun Wang, Yukun Zhu, Ruoming Pang, Vijay Vasudevan, et~al.
\newblock Searching for mobilenetv3.
\newblock In {\em ICCV}, 2019.

\bibitem{mnetv1}
Andrew~G Howard, Menglong Zhu, Bo~Chen, Dmitry Kalenichenko, Weijun Wang, Tobias Weyand, Marco Andreetto, and Hartwig Adam.
\newblock Mobilenets: Efficient convolutional neural networks for mobile vision applications.
\newblock {\em arXiv preprint arXiv:1704.04861}, 2017.

\bibitem{hu2023you}
Jie Hu, Linyan Huang, Tianhe Ren, Shengchuan Zhang, Rongrong Ji, and Liujuan Cao.
\newblock You only segment once: Towards real-time panoptic segmentation.
\newblock In {\em CVPR}, 2023.

\bibitem{huang2021fapn}
Shihua Huang, Zhichao Lu, Ran Cheng, and Cheng He.
\newblock Fapn: Feature-aligned {Pyramid} {Network} for {Dense} {Image} {Prediction}.
\newblock In {\em ICCV}, 2021.

\bibitem{squeezenet}
Forrest~N Iandola, Song Han, Matthew~W Moskewicz, Khalid Ashraf, William~J Dally, and Kurt Keutzer.
\newblock Squeezenet: Alexnet-level accuracy with 50x fewer parameters and< 0.5 mb model size.
\newblock {\em arXiv preprint arXiv:1602.07360}, 2016.

\bibitem{transfiner}
Lei Ke, Martin Danelljan, Xia Li, Yu-Wing Tai, Chi-Keung Tang, and Fisher Yu.
\newblock Mask transfiner for high-quality instance segmentation.
\newblock In {\em CVPR}, 2022.

\bibitem{sam_hq}
Lei Ke, Mingqiao Ye, Martin Danelljan, Yifan Liu, Yu-Wing Tai, Chi-Keung Tang, and Fisher Yu.
\newblock Segment anything in high quality.
\newblock In {\em NeurIPS}, 2023.

\bibitem{kirillov2023segment}
Alexander Kirillov, Eric Mintun, Nikhila Ravi, Hanzi Mao, Chloe Rolland, Laura Gustafson, Tete Xiao, Spencer Whitehead, Alexander~C Berg, Wan-Yen Lo, et~al.
\newblock Segment anything.
\newblock {\em ICCV}, 2023.

\bibitem{kirillov2020pointrend}
Alexander Kirillov, Yuxin Wu, Kaiming He, and Ross Girshick.
\newblock Pointrend: Image segmentation as rendering.
\newblock In {\em CVPR}, 2020.

\bibitem{nextvit}
Jiashi Li, Xin Xia, Wei Li, Huixia Li, Xing Wang, Xuefeng Xiao, Rui Wang, Min Zheng, and Xin Pan.
\newblock Next-vit: Next generation vision transformer for efficient deployment in realistic industrial scenarios.
\newblock {\em arXiv preprint arXiv:2207.05501}, 2022.

\bibitem{li2024videomamba}
Kunchang Li, Xinhao Li, Yi~Wang, Yinan He, Yali Wang, Limin Wang, and Yu~Qiao.
\newblock Videomamba: State space model for efficient video understanding.
\newblock {\em arXiv}, 2024.

\bibitem{li2023transformer}
Xiangtai Li, Henghui Ding, Wenwei Zhang, Haobo Yuan, Guangliang Cheng, Pang Jiangmiao, Kai Chen, Ziwei Liu, and Chen~Change Loy.
\newblock Transformer-based visual segmentation: A survey.
\newblock {\em arXiv pre-print}, 2023.

\bibitem{li2020improving}
Xiangtai Li, Xia Li, Li~Zhang, Guangliang Cheng, Jianping Shi, Zhouchen Lin, Shaohua Tan, and Yunhai Tong.
\newblock Improving semantic segmentation via decoupled body and edge supervision.
\newblock In {\em ECCV}, 2020.

\bibitem{SFnet}
Xiangtai Li, Ansheng You, Zeping Zhu, Houlong Zhao, Maoke Yang, Kuiyuan Yang, and Yunhai Tong.
\newblock Semantic flow for fast and accurate scene parsing.
\newblock In {\em ECCV}, 2020.

\bibitem{Li2022ExploringPV}
Yanghao Li, Hanzi Mao, Ross~B. Girshick, and Kaiming He.
\newblock Exploring plain vision transformer backbones for object detection.
\newblock In {\em ECCV}, 2022.

\bibitem{liew2021deep}
Jun~Hao Liew, Scott Cohen, Brian Price, Long Mai, and Jiashi Feng.
\newblock Deep interactive thin object selection.
\newblock In {\em WACV}, 2021.

\bibitem{coco_dataset}
Tsung-Yi Lin, Michael Maire, Serge Belongie, James Hays, Pietro Perona, Deva Ramanan, Piotr Doll{\'a}r, and C~Lawrence Zitnick.
\newblock Microsoft coco: Common objects in context.
\newblock In {\em ECCV}, 2014.

\bibitem{liu2024rethinking}
Qin Liu, Jaemin Cho, Mohit Bansal, and Marc Niethammer.
\newblock Rethinking interactive image segmentation with low latency, high quality, and diverse prompts.
\newblock In {\em CVPR}, 2024.

\bibitem{liu2024vmamba}
Yue Liu, Yunjie Tian, Yuzhong Zhao, Hongtian Yu, Lingxi Xie, Yaowei Wang, Qixiang Ye, and Yunfan Liu.
\newblock Vmamba: Visual state space model.
\newblock {\em arXiv preprint arXiv:2401.10166}, 2024.

\bibitem{liu2021swin}
Ze~Liu, Yutong Lin, Yue Cao, Han Hu, Yixuan Wei, Zheng Zhang, Stephen Lin, and Baining Guo.
\newblock Swin transformer: Hierarchical vision transformer using shifted windows.
\newblock {\em ICCV}, 2021.

\bibitem{ADAMW}
Ilya Loshchilov and Frank Hutter.
\newblock Decoupled weight decay regularization.
\newblock In {\em ICLR}, 2019.

\bibitem{mocovit}
Hailong Ma, Xin Xia, Xing Wang, Xuefeng Xiao, Jiashi Li, and Min Zheng.
\newblock Mocovit: Mobile convolutional vision transformer.
\newblock {\em arXiv preprint arXiv:2205.12635}, 2022.

\bibitem{Ma2024SegmentAI}
Jun Ma, Yuting He, Feifei Li, Lin Han, Chenyu You, and Bo~Wang.
\newblock Segment anything in medical images.
\newblock {\em Nature Communications}, 2024.

\bibitem{shufflenetv2}
Ningning Ma, Xiangyu Zhang, Hai-Tao Zheng, and Jian Sun.
\newblock Shufflenet v2: Practical guidelines for efficient cnn architecture design.
\newblock In {\em ECCV}, 2018.

\bibitem{edgenext}
Muhammad Maaz, Abdelrahman Shaker, Hisham Cholakkal, Salman Khan, Syed~Waqas Zamir, Rao~Muhammad Anwer, and Fahad~Shahbaz Khan.
\newblock Edgenext: efficiently amalgamated cnn-transformer architecture for mobile vision applications.
\newblock In {\em ECCVW}, 2022.

\bibitem{mvitv1}
Sachin Mehta and Mohammad Rastegari.
\newblock Mobilevit: Light-weight, general-purpose, and mobile-friendly vision transformer.
\newblock In {\em ICLR}, 2022.

\bibitem{mehta2018espnet}
Sachin Mehta, Mohammad Rastegari, Anat Caspi, Linda Shapiro, and Hannaneh Hajishirzi.
\newblock Espnet: Efficient spatial pyramid of dilated convolutions for semantic segmentation.
\newblock In {\em ECCV}, 2018.

\bibitem{espnetv2}
Sachin Mehta, Mohammad Rastegari, Linda Shapiro, and Hannaneh Hajishirzi.
\newblock Espnetv2: A light-weight, power efficient, and general purpose convolutional neural network.
\newblock In {\em CVPR}, 2019.

\bibitem{milletari2016v}
Fausto Milletari, Nassir Navab, and Seyed-Ahmad Ahmadi.
\newblock V-net: Fully convolutional neural networks for volumetric medical image segmentation.
\newblock In {\em 3DV}, 2016.

\bibitem{mohan2020efficientps}
Rohit Mohan and Abhinav Valada.
\newblock Efficientps: Efficient panoptic segmentation.
\newblock {\em IJCV}, 2021.

\bibitem{edgevit}
Junting Pan, Adrian Bulat, Fuwen Tan, Xiatian Zhu, Lukasz Dudziak, Hongsheng Li, Georgios Tzimiropoulos, and Brais Martinez.
\newblock Edgevits: Competing light-weight cnns on mobile devices with vision transformers.
\newblock {\em ECCV}, 2022.

\bibitem{peng2023rwkv}
Bo~Peng, Eric Alcaide, Quentin Anthony, Alon Albalak, Samuel Arcadinho, Huanqi Cao, Xin Cheng, Michael Chung, Matteo Grella, Kranthi~Kiran GV, et~al.
\newblock Rwkv: Reinventing rnns for the transformer era.
\newblock {\em arXiv preprint arXiv:2305.13048}, 2023.

\bibitem{peng2024eagle}
Bo~Peng, Daniel Goldstein, Quentin Anthony, Alon Albalak, Eric Alcaide, Stella Biderman, Eugene Cheah, Teddy Ferdinan, Haowen Hou, Przemys{\l}aw Kazienko, et~al.
\newblock Eagle and finch: Rwkv with matrix-valued states and dynamic recurrence.
\newblock {\em arXiv preprint arXiv:2404.05892}, 2024.

\bibitem{qi2022fine}
Lu~Qi, Jason Kuen, Tiancheng Shen, Jiuxiang Gu, Weidong Guo, Jiaya Jia, Zhe Lin, and Ming-Hsuan Yang.
\newblock High-quality entity segmentation.
\newblock In {\em ICCV}, 2023.

\bibitem{qi2022open}
Lu~Qi, Jason Kuen, Yi~Wang, Jiuxiang Gu, Hengshuang Zhao, Philip Torr, Zhe Lin, and Jiaya Jia.
\newblock Open world entity segmentation.
\newblock {\em TPAMI}, 2022.

\bibitem{qin2022}
Xuebin Qin, Hang Dai, Xiaobin Hu, Deng-Ping Fan, Ling Shao, and Luc~Van Gool.
\newblock Highly accurate dichotomous image segmentation.
\newblock In {\em ECCV}, 2022.

\bibitem{qin2024hgrn2}
Zhen Qin, Songlin Yang, Weixuan Sun, Xuyang Shen, Dong Li, Weigao Sun, and Yiran Zhong.
\newblock Hgrn2: Gated linear rnns with state expansion.
\newblock {\em arXiv preprint arXiv:2404.07904}, 2024.

\bibitem{qin2024hierarchically}
Zhen Qin, Songlin Yang, and Yiran Zhong.
\newblock Hierarchically gated recurrent neural network for sequence modeling.
\newblock {\em NeurIPS}, 2024.

\bibitem{russakovsky2015imagenet}
Olga Russakovsky, Jia Deng, Hao Su, Jonathan Krause, Sanjeev Satheesh, Sean Ma, Zhiheng Huang, Andrej Karpathy, Aditya Khosla, Michael Bernstein, et~al.
\newblock Imagenet large scale visual recognition challenge.
\newblock {\em IJCV}, 2015.

\bibitem{mnetv2}
Mark Sandler, Andrew Howard, Menglong Zhu, Andrey Zhmoginov, and Liang-Chieh Chen.
\newblock Mobilenetv2: Inverted residuals and linear bottlenecks.
\newblock In {\em CVPR}, 2018.

\bibitem{shen2022high}
Tiancheng Shen, Yuechen Zhang, Lu~Qi, Jason Kuen, Xingyu Xie, Jianlong Wu, Zhe Lin, and Jiaya Jia.
\newblock High quality segmentation for ultra high-resolution images.
\newblock In {\em CVPR}, 2022.

\bibitem{shen2021efficient}
Zhuoran Shen, Mingyuan Zhang, Haiyu Zhao, Shuai Yi, and Hongsheng Li.
\newblock Efficient attention: Attention with linear complexities.
\newblock In {\em WACV}, 2021.

\bibitem{song2024ba}
Yiran Song, Qianyu Zhou, Xiangtai Li, Deng-Ping Fan, Xuequan Lu, and Lizhuang Ma.
\newblock Ba-sam: Scalable bias-mode attention mask for segment anything model.
\newblock {\em CVPR}, 2024.

\bibitem{sun2023remax}
Shuyang Sun, Weijun Wang, Qihang Yu, Andrew Howard, Philip Torr, and Liang-Chieh Chen.
\newblock Remax: Relaxing for better training on efficient panoptic segmentation.
\newblock In {\em NeurIPS}, 2023.

\bibitem{wan2023seaformer}
Qiang Wan, Zilong Huang, Jiachen Lu, Gang Yu, and Li~Zhang.
\newblock Seaformer: Squeeze-enhanced axial transformer for mobile semantic segmentation.
\newblock In {\em ICLR}, 2023.

\bibitem{SegRefiner}
Mengyu Wang, Henghui Ding, Jun~Hao Liew, Jiajun Liu, Yao Zhao, and Yunchao Wei.
\newblock {SegRefiner}: Towards model-agnostic segmentation refinement with discrete diffusion process.
\newblock In {\em NeurIPS}, 2023.

\bibitem{xiao2018unified}
Tete Xiao, Yingcheng Liu, Bolei Zhou, Yuning Jiang, and Jian Sun.
\newblock Unified perceptual parsing for scene understanding.
\newblock In {\em ECCV}, 2018.

\bibitem{xiong2023efficientsam}
Yunyang Xiong, Bala Varadarajan, Lemeng Wu, Xiaoyu Xiang, Fanyi Xiao, Chenchen Zhu, Xiaoliang Dai, Dilin Wang, Fei Sun, Forrest Iandola, et~al.
\newblock Efficientsam: Leveraged masked image pretraining for efficient segment anything.
\newblock In {\em CVPR}, 2024.

\bibitem{xu2024rapsam}
Shilin Xu, Haobo Yuan, Qingyu Shi, Lu~Qi, Jingbo Wang, Yibo Yang, Yining Li, Kai Chen, Yunhai Tong, Bernard Ghanem, Xiangtai Li, and Ming-Hsuan Yang.
\newblock Rap-sam:towards real-time all-purpose segment anything.
\newblock {\em arXiv preprint}, 2024.

\bibitem{CrossVIS}
Shusheng Yang, Yuxin Fang, Xinggang Wang, Yu~Li, Chen Fang, Ying Shan, Bin Feng, and Wenyu Liu.
\newblock Crossover learning for fast online video instance segmentation.
\newblock In {\em ICCV}, 2021.

\bibitem{yang2023gated}
Songlin Yang, Bailin Wang, Yikang Shen, Rameswar Panda, and Yoon Kim.
\newblock Gated linear attention transformers with hardware-efficient training.
\newblock {\em arXiv preprint arXiv:2312.06635}, 2023.

\bibitem{yu2021bisenetv2}
Changqian Yu, Changxin Gao, Jingbo Wang, Gang Yu, Chunhua Shen, and Nong Sang.
\newblock Bisenet v2: Bilateral network with guided aggregation for real-time semantic segmentation.
\newblock {\em IJCV}, 2021.

\bibitem{bisenet}
Changqian Yu, Jingbo Wang, Chao Peng, Changxin Gao, Gang Yu, and Nong Sang.
\newblock Bisenet: Bilateral segmentation network for real-time semantic segmentation.
\newblock In {\em ECCV}, 2018.

\bibitem{yuan2024ovsam}
Haobo Yuan, Xiangtai Li, Chong Zhou, Yining Li, Kai Chen, and Chen~Change Loy.
\newblock Open-vocabulary sam: Segment and recognize twenty-thousand classes interactively.
\newblock {\em arXiv preprint}, 2024.

\bibitem{Zeng2019TowardsHS}
Yi~Zeng, Pingping Zhang, Jianming Zhang, Zhe~L. Lin, and Huchuan Lu.
\newblock Towards high-resolution salient object detection.
\newblock In {\em ICCV}, 2019.

\bibitem{mobile_sam}
Chaoning Zhang, Dongshen Han, Yu~Qiao, Jung~Uk Kim, Sung-Ho Bae, Seungkyu Lee, and Choong~Seon Hong.
\newblock Faster segment anything: Towards lightweight sam for mobile applications.
\newblock {\em arXiv preprint arXiv:2306.14289}, 2023.

\bibitem{EMOiccv23}
Jiangning Zhang, Xiangtai Li, Jian Li, Liang Liu, Zhucun Xue, Boshen Zhang, Zhengkai Jiang, Tianxin Huang, Yabiao Wang, and Chengjie Wang.
\newblock Rethinking mobile block for efficient attention-based models.
\newblock In {\em ICCV}, 2023.

\bibitem{zhang2023mobileinst}
Renhong Zhang, Tianheng Cheng, Shusheng Yang, Haoyi Jiang, Shuai Zhang, Jiancheng Lyu, Xin Li, Xiaowen Ying, Dashan Gao, Wenyu Liu, et~al.
\newblock Mobileinst: Video instance segmentation on the mobile.
\newblock {\em AAAI}, 2024.

\bibitem{zhang2024point}
Tao Zhang, Xiangtai Li, Haobo Yuan, Shunping Ji, and Shuicheng Yan.
\newblock Point cloud mamba: Point cloud learning via state space model.
\newblock {\em arXiv preprint arXiv:2403.00762}, 2024.

\bibitem{zhang2022topformer}
Wenqiang Zhang, Zilong Huang, Guozhong Luo, Tao Chen, Xinggang Wang, Wenyu Liu, Gang Yu, and Chunhua Shen.
\newblock Topformer: Token pyramid transformer for mobile semantic segmentation.
\newblock In {\em CVPR}, 2022.

\bibitem{shufflenetv1}
Xiangyu Zhang, Xinyu Zhou, Mengxiao Lin, and Jian Sun.
\newblock Shufflenet: An extremely efficient convolutional neural network for mobile devices.
\newblock In {\em CVPR}, 2018.

\bibitem{ICnet}
Hengshuang Zhao, Xiaojuan Qi, Xiaoyong Shen, Jianping Shi, and Jiaya Jia.
\newblock Icnet for real-time semantic segmentation on high-resolution images.
\newblock In {\em ECCV}, 2018.

\bibitem{zhao2023fastsam}
Xu~Zhao, Wenchao Ding, Yongqi An, Yinglong Du, Tao Yu, Min Li, Ming Tang, and Jinqiao Wang.
\newblock Fast segment anything.
\newblock {\em arXiv}, 2023.

\bibitem{ADE20K}
Bolei Zhou, Hang Zhao, Xavier Puig, Sanja Fidler, Adela Barriuso, and Antonio Torralba.
\newblock {Semantic understanding of scenes through the ADE20K dataset}.
\newblock {\em CVPR}, 2017.

\bibitem{zhou2023edgesam}
Chong Zhou, Xiangtai Li, Chen~Change Loy, and Bo~Dai.
\newblock Edgesam: Prompt-in-the-loop distillation for on-device deployment of sam.
\newblock {\em arXiv preprint arXiv:2312.06660}, 2023.

\bibitem{zhou2023context}
Qianyu Zhou, Zhengyang Feng, Qiqi Gu, Jiangmiao Pang, Guangliang Cheng, Xuequan Lu, Jianping Shi, and Lizhuang Ma.
\newblock Context-aware mixup for domain adaptive semantic segmentation.
\newblock {\em IEEE TCSVT}, 2023.

\bibitem{zhou2022transvod}
Qianyu Zhou, Xiangtai Li, Lu~He, Yibo Yang, Guangliang Cheng, Yunhai Tong, Lizhuang Ma, and Dacheng Tao.
\newblock Transvod: End-to-end video object detection with spatial-temporal transformers.
\newblock {\em TPAMI}, 2022.

\bibitem{zhu2024vision}
Lianghui Zhu, Bencheng Liao, Qian Zhang, Xinlong Wang, Wenyu Liu, and Xinggang Wang.
\newblock Vision mamba: Efficient visual representation learning with bidirectional state space model.
\newblock {\em arXiv preprint arXiv:2401.09417}, 2024.

\end{thebibliography}
}
\newpage
\appendix

\noindent
\textbf{Overview.}
In the appendix, we present more implementation details (Section~\ref{sec:appendix_impl}), more visualization results (Section~\ref{sec:appendix_vis}), more ablation studies (Section~\ref{sec:appendix_com}) and limitations of our work (Section~\ref{sec:appendix_limit}).
\section{More Implementation Details.}\label{sec:appendix_impl}
\begin{figure}
    \centering
    \includegraphics[width=\linewidth]{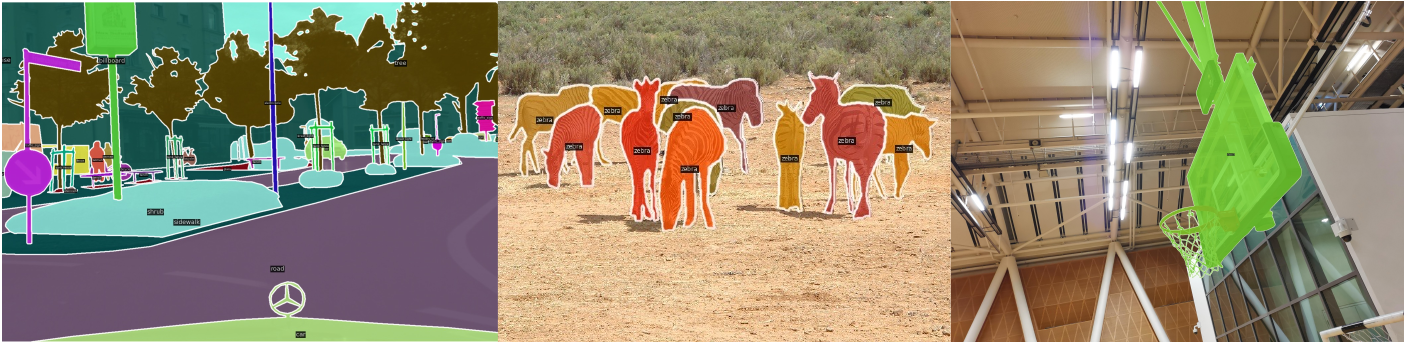}
    \caption{The training datasets of RWKV-SAM. (Left) EntitySeg~\cite{qi2022fine} dataset. (Middle) COCONut-B~\cite{deng2024coconut} dataset. (Right) DIS5K~\cite{qin2022} dataset.}
    \label{fig:datasests}
\end{figure}
\noindent
\textbf{Training Datasets.} We visualize the training datasets of RWKV-SAM in Figure~\ref{fig:datasests}. EntitySeg~\cite{qi2022fine} dataset has the most diversified scenes and provides detailed annotations for each entity. COCONut-B~\cite{deng2024coconut} relabeled COCO~\cite{coco_dataset} and provides finer annotations. DIS5K~\cite{qin2022} dataset provides single-object high-quality annotation. The annotations are object-centric and are large in the image.
\begin{figure}
    \centering
    \includegraphics[width=\linewidth]{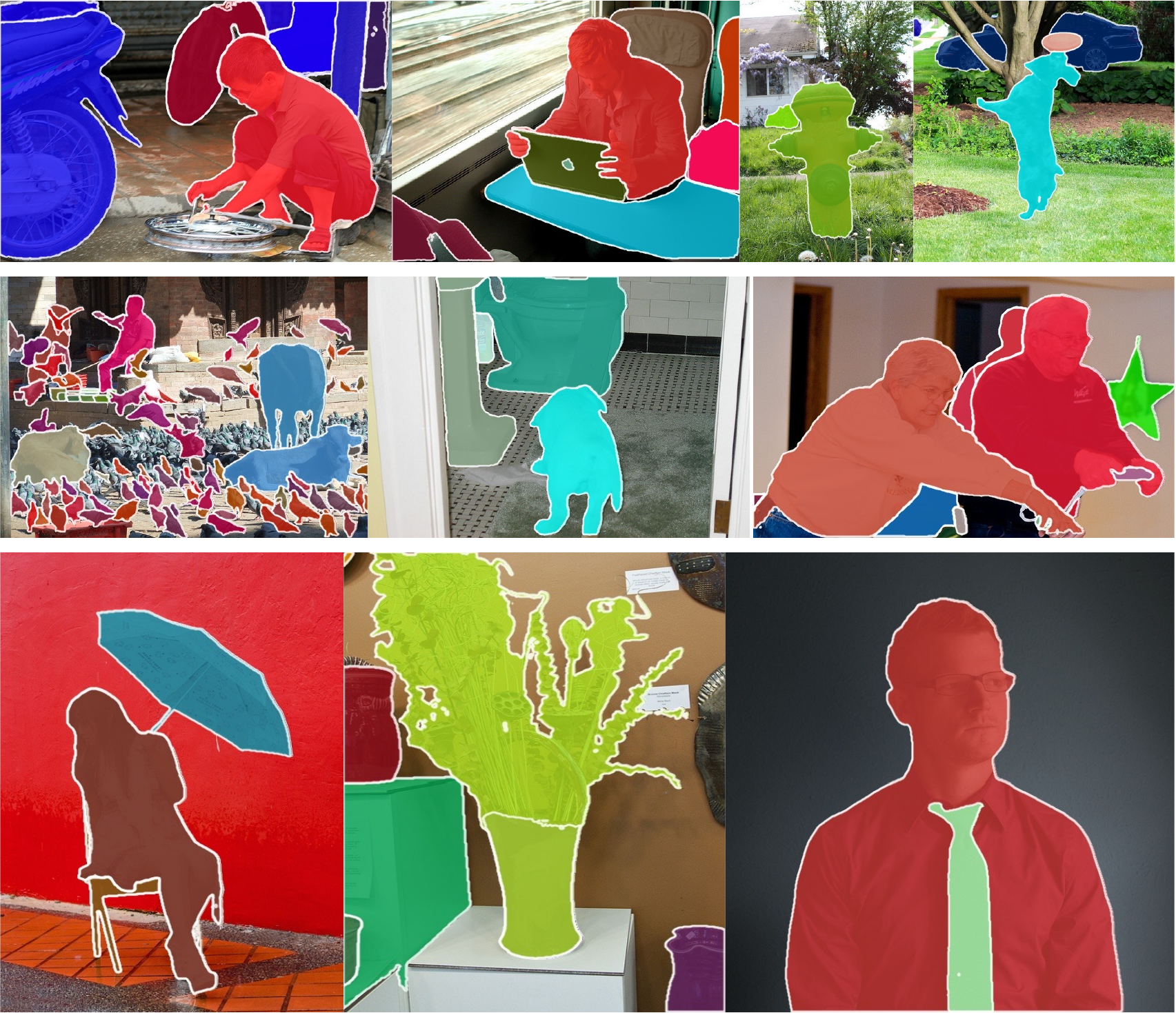}
    \caption{More visualization results on COCO dataset of RWKV-SAM.}
    \label{fig:coco}
\end{figure}
\noindent
\textbf{Training Details.}
In Section~\ref{sec:exp_1}, we already mention our training involves two sessions. In the first session on the SA-1B dataset, we set the learning rate to 0.0001 with the AdamW~\cite{ADAMW} optimizer. We use cosine annealing for the learning rate schedule for 24 epochs. In the second session, we also set the learning rate to 0.0001 with the AdamW~\cite{ADAMW} optimizer and use cosine annealing for the learning rate schedule for 6 epochs. In each session, the total number of training samples is roughly equal to 24 times the COCO training dataset. The batch size in each session is 32.

The training hyperparameters of the pretraining session on the ImageNet-1k dataset mainly follow Swin-Transformer~\cite{liu2021swin}, with 0.001 learning rate, AdamW optimizer, and 1024 batch size. But we train our RWKV-SAM for 120 epochs to save the computational cost. The training on ADE20K~\cite{ADE20K} dataset takes 160k steps with a batch size 16. We do not use any test-time augmentation for fair comparison.

\noindent
\textbf{Training time.}
We train our RWKV-SAM model on 16 A100 GPUs. The first session takes about 5 hours, and the second session takes about 16 hours.

\section{More Visualization Results}\label{sec:appendix_vis}
\noindent
\textbf{More Visualization Results on COCO dataset.}
We present more visualization results on COCO datasets. As demonstrated in Figure~\ref{fig:coco}, our RWKV-SAM can segment objects in high quality. The results show that our RWKV-SAM can segment various objects, even in complex scenes.

\begin{figure}
    \centering
    \includegraphics[width=\linewidth]{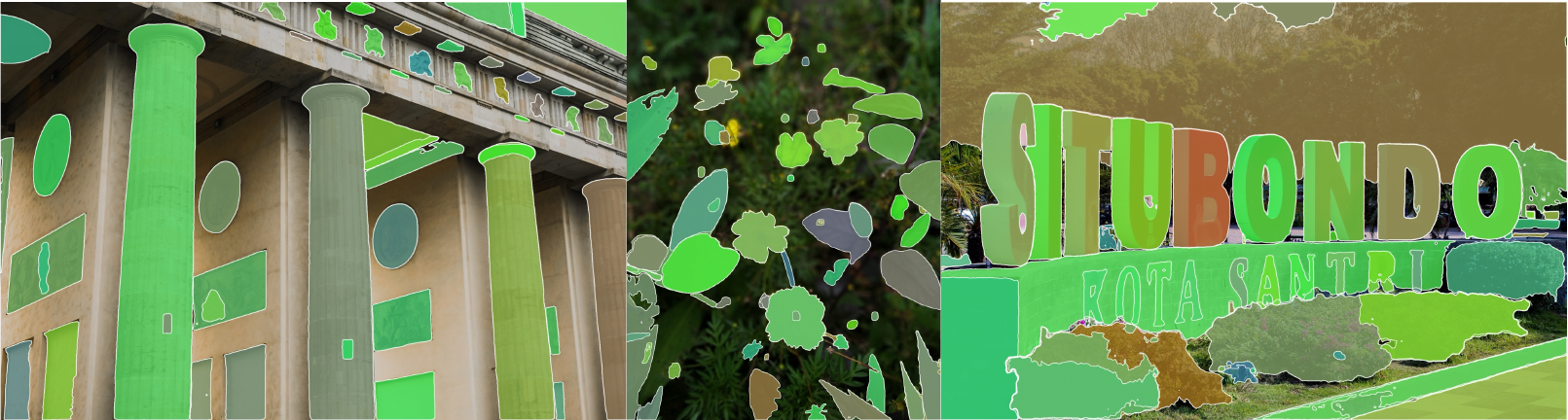}
    \caption{More visualization results on SA-1B dataset of RWKV-SAM.}
    \label{fig:sam}
\end{figure}
\noindent
\textbf{More Visualization Results on SA-1B datasets.}
We present more visualization results on the SA-1B dataset to demonstrate the performance in the open world. Note that our training data do not have the SA-1B dataset in the second training session. As shown in Figure~\ref{fig:sam}, it can provide surprising high-quality segmentation results with good details on SA-1B.

\section{More Ablation Study.}\label{sec:appendix_com}
\begin{table}
\centering
 \renewcommand\tabcolsep{2pt}
 \renewcommand\arraystretch{1.2}
  \resizebox{0.42\linewidth}{!}{
     \begin{tabular}{l|ccc} 
       \toprule
        Method  & DIS & COIFT & HRSOD\\
        \midrule
        RWKV-SAM & 80.5  & 94.1& 92.5\\
        RWKV-SAM (no dist) & 78.3 & 93.4 & 92.3\\
        \bottomrule
     \end{tabular}
 }
 \vspace{1mm}
 \caption{Ablation Study on the backbone distillation.}
 \label{tab:ablation_dist}
\end{table}
We do the ablation study on the backbone distillation for our RWKV-SAM model. The results are shown in Table~\ref{tab:ablation_dist}. Without the backbone distillation, the RWKV-SAM cannot get the existing knowledge from the SAM encoder well and thus performs poorly. 

\begin{figure}
    \centering
    \includegraphics[width=0.8\linewidth]{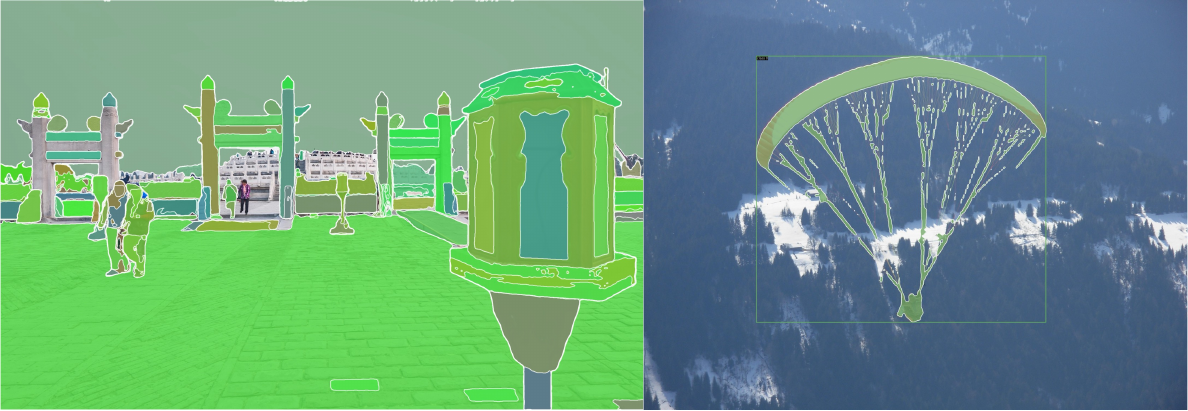}
    \caption{Failure cases of RWKV-SAM.}
    \label{fig:fail}
\end{figure}

\section{Limitation and Future Work}\label{sec:appendix_limit}
\noindent
\textbf{Limitation.} Our RWKV-SAM is a SAM-like prompt-based segmentation method, which means our method cannot propose and recognize objects well like instance segmentation methods such as Mask2Former~\cite{cheng2021mask2former}. Our RWKV-SAM also falls short in some part-level segmentation or very thin objects (such as wire). We provide some failure cases in Figure~\ref{fig:fail}.

\noindent
\textbf{Future work.}
While our RWKV-SAM already has high-quality segmentation results, it may fail in some cases (e.g., failure cases in Figure~\ref{fig:fail}). Future work may incorporate more training datasets to support more complex scenarios. We aim to continue exploring this direction. For example, we aim to adopt the full SA-1B dataset for co-training. 

\textbf{Broader Impact.}
Our method provides an efficient and high-quality interactive segmentation tool, which may enable downstream tasks such as image editing. We do not think it will bring any additional negative social impact compared to SAM~\cite{kirillov2023segment}.

\end{document}